\documentclass{article}

\PassOptionsToPackage{numbers,sort&compress}{natbib}
\usepackage[dblblindworkshop, final]{neurips_2025}

\usepackage[utf8]{inputenc} 
\usepackage[T1]{fontenc}   
\usepackage[colorlinks,linkcolor=blue,citecolor=blue,urlcolor=blue]{hyperref}    
\usepackage{url}           
\usepackage{booktabs}       
\usepackage{amsfonts}       
\usepackage{nicefrac}      
\usepackage{microtype}      
\usepackage{xcolor}        

\usepackage{amsmath,amssymb,mathtools}
\usepackage{algorithm}
\usepackage{algorithmic}
\usepackage{multirow}
\usepackage{etoolbox}

\newcommand{\email}[1]{%
  {\hypersetup{urlcolor=black}\href{mailto:#1}{\begingroup\ttfamily #1\endgroup}}%
}

\workshoptitle{Efficient Reasoning}

\usepackage{xspace}
\IfFileExists{relsize.sty}{\usepackage{relsize}}{} 

\DeclareRobustCommand{\leash}{%
  \begingroup
    \IfFileExists{relsize.sty}{\relsize{-0.5}}{\small} 
    \textbf{\textsc{LEASH}}%
  \endgroup\xspace}

\title{Logit–Entropy Adaptive Stopping Heuristic for Efficient Chain-of-Thought Reasoning}

\author{%
  Mohammad Atif Quamar \\
  Independent Researcher\\
  \email{atif7102@gmail.com} \\
  \And
  Mohammad Areeb \\
  Purdue University\\
  \email{mareeb@purdue.edu} \\
}

\begin{document}

\maketitle

\begin{abstract}
Chain-of-Thought (CoT) prompting is a key technique for enabling complex reasoning in large language models. However, generating full, fixed-length rationales is computationally wasteful, inflating both token usage and latency. We introduce \leash: \textbf{\underline{L}ogit-\underline{E}ntropy \underline{A}daptive \underline{S}topping \underline{H}euristic}, a training-free decoding algorithm that adaptively halts rationale generation. \leash{} monitors two intrinsic signals: the slope of token-level entropy and the improvement in the top-logit margin. It terminates the generation once both signals plateau, indicating the model has reached a stable reasoning state. Across four instruction-tuned models on the GSM8K and AQuA-RAT benchmarks, \leash{} reduces average token generation by $\approx$ 30--35\% and latency by $\approx$ 27\%, while incurring a $\approx$ 10 p.p. accuracy drop relative to CoT. \leash is model-agnostic and requires no additional training or supervision, offering a simple and efficient alternative to CoT decoding.
\end{abstract}

\section{Introduction}

Large language models solve many reasoning problems more reliably when prompted to “think out loud” using chain-of-thought (CoT) decoding~\citep{wei2023chainofthoughtpromptingelicitsreasoning}. Yet those rationales are costly: Vanilla CoT and vote-heavy schemes inflate token usage and tail latency, which limits deployment under tight budgets and interactive constraints. The core challenge is to decide, online and per question, when enough reasoning has been generated, early enough to save tokens, but not so early that accuracy suffers.

Existing approaches offer unsatisfying trade-offs. Fixed budgets ignore instance difficulty and routinely over-generate~\citep{han2024token}. Heuristic triggers (e.g., stopping on “Therefore,” or punctuation patterns) are brittle and prompt-dependent~\citep{kojima2022zeroshotcot}. Multi-sample reranking improves quality but wastes compute on full sequences that were already off-trajectory~\citep{wang2022selfconsistency}. What is missing is a training-free, model-agnostic criterion that uses signals already available at decoding time to adaptively halt reasoning without extra models, supervision, or architectural changes.

We introduce \textbf{Logit–Entropy Adaptive Stopping Heuristic (\leash)}, a simple decoding-time algorithm that monitors two intrinsic indicators of reasoning convergence: the local slope of token-level entropy and the improvement in the top-logit margin. \leash generates a brief rationale and halts when both signals plateau within a short sliding window after a small minimum length, then elicits a concise final answer. Because \leash relies only on logits already produced by the base model, it is gradient-free, drop-in, and compatible with greedy or sampled decoding, quantized or full-precision inference, and common toolchains.

Prior efforts to curb over-generation in CoT either truncate at a fixed depth, add auxiliary heads or verifiers, or stop on a \emph{level} signal such as an absolute entropy threshold over an explicit answer set~\citep{laaouach2025haltcot}. In contrast, we use intrinsic token-level signals available under standard top-\(p\) sampling: the windowed slope of next-token entropy and the trend in the top-logit margin, together with a \(p_{\max}\) saturation guard, to detect convergence. As our approach operates on logits already computed during decoding, it integrates with existing APIs, supports quantized inference, and extends to free-form numeric reasoning without additional scaffolding. Empirically, this instance-wise token-granular rule yields large reductions in generated tokens and latency while maintaining competitive accuracy, offering a training-free alternative to answer-entropy halting.

Our empirical study focuses on grade-school math, where CoT is standard. On GSM8K~\citep{cobbe2021gsm8k} with four instruction-tuned models, \leash retains \(\approx 85\%\) of vanilla CoT accuracy, while using \(\sim 30-35\%\) fewer tokens and cutting end-to-end latency by \(\sim 25-30\%\).

To probe the generality of our findings, we report results on the test split of the AQuA-RAT dataset~\citep{ling2017program}. On this benchmark, \leash closely tracks the accuracy of vanilla-CoT while reducing compute. We also examine robustness across sampling temperatures and decoding settings, finding that \leash maintains performance without per-task retuning. Taken together, these results position \leash as a practical, training-free alternative to vanilla CoT: it adapts rationale length to item difficulty, preserves accuracy under tight budgets, and delivers immediate token and latency savings for real-world deployments.

\section{LEASH: Training-Free Stopping for Chain of Thought}
\label{sec:method}

We consider an instruction-tuned language model that first generates a chain-of-thought (CoT) rationale and then a short final answer. For a prompt $x$, let $y_{1:t}$ denote the partial rationale at step $t$, and let $z_t\in\mathbb{R}^{V}$ be the next-token logits over a vocabulary of size $V$.

During rationale generation, end-of-sequence termination is disabled and halting is governed by the rule in \eqref{eq:stopping}. After halting, a second prompt requests a short final answer.

For Numerical stability, we upcast logits to \texttt{fp32}, replace non-finite entries with zero, and clip componentwise logits to a fixed band $[-B, B]$, i.e., $\tilde z_t \;=\; \mathrm{clip}\!\left(\mathrm{finite}(z_t),\, -B,\, B\right)$.

Let $p_t = \mathrm{softmax}(\tilde z_t)$ denote the token probabilities and $\ell_t = \mathrm{logsoftmax}(\tilde z_t)$ be the log-probabilities at step $t$. We monitor two primary intrinsic signals: the token-level entropy $H_t$ and the top-two log-probability margin $M_t$. Entropy measures the models uncertainty, while the margin measures its confidence in the top choice.
\begin{equation}
H_t \;=\; -\sum_{v=1}^{V} p_t(v)\,\log p_t(v)
\label{eq:entropy}
\end{equation}
\begin{equation}
M_t \;=\; \ell_t^{(1)}-\ell_t^{(2)}
\label{eq:margin-logprob}
\end{equation}
where $\ell_t^{(1)}$ and $\ell_t^{(2)}$ are the log-probabilities of the top-two tokens. While $M_t$ is algebraically equivalent to the logit-space margin $z_t^{(1)}-z_t^{(2)}$, we compute it using log-probabilities for numerical stability.
Highly confident steps, where the peak probability $p_{\max}(t) = \max_{v} p_t(v)$ exceeds a threshold $\tau_p$, are treated as "saturated" and are excluded from our trend analysis. We use an indicator $\Sigma_t$ to mark these steps:
\begin{equation}
\Sigma_t \;=\; \mathbb{I}\!\left[p_{\max}(t)\ \ge\ \tau_p\right]
\label{eq:saturation}
\end{equation}
\paragraph{Windowed trends.}
Given a window size $k\ge1$, we compute the $k$-step entropy slope $s_H$ and the $k$-step margin improvement $\Delta M$ for all non-saturated steps.

\begin{equation}
s_H(t;k) \;=\; \frac{H_t - H_{t-k}}{k}
\label{eq:entropy-slope}
\end{equation}
\begin{equation}
\Delta M(t;k) \;=\; M_t - M_{t-k}
\label{eq:margin-improve}
\end{equation}

\paragraph{Adaptive Stopping Criterion.}
Our stopping rule is determined by a per-step plateau test, $\Pi_t$, which is active only for non-saturated steps ($\Sigma_t=0$). The test passes if the entropy slope has flattened and the margin improvement has stalled, given small tolerances $\varepsilon_H > 0$ and $\delta_M > 0$:

\begin{equation}
\Pi_t \;=\; \mathbb{I}\!\big[\, s_H(t;k)\ \ge\ -\varepsilon_H \,\big]\ \cdot\ \mathbb{I}\!\big[\, \Delta M(t;k)\ \le\ \delta_M \,\big]\ \cdot\ \mathbb{I}\!\big[\, \Sigma_t=0 \,\big]
\label{eq:plateau}
\end{equation}

The final stopping time $\tau$ is the first step $t$ that satisfies three conditions:
(i) It is past a minimum warm-up period, $t_{\min} = \max(m+w, k+L)$.
(ii) A majority of the last $L$ non-saturated steps (indexed by $\mathcal{J}_L(t)$) have passed the plateau test.
(iii) An entropy-drop gate, $H_{\mathrm{ref}} - H_t \ge \gamma$, is satisfied, preventing premature stops. $H_{\mathrm{ref}} = \mathrm{median}(H_1, \dots, H_k)$ is a reference entropy computed over the first $k$ steps.
The full stopping criterion, capped at a maximum length $M$, is:

\begin{equation}
\tau \;=\; \min\!\left\{\, t \ \ge\ t_{\min}\ :\ 
\sum_{j \in \mathcal{J}_L(t)} \Pi_j \ \ge\ \left\lceil \frac{|\mathcal{J}_L(t)|}{2} \right\rceil
\ \land\ \big(H_{\mathrm{ref}} - H_t \ge \gamma\big)
\right\} \ \wedge\ M
\label{eq:stopping}
\end{equation}

The rationale stage disables EOS, so halting is governed by \eqref{eq:stopping}. After stopping at $\tau$, the model is prompted again to emit only the short final answer. Our complete method is given in Algorithm~\ref{alg:esa}.

\paragraph{Implementation Details.}
The signals in \eqref{eq:entropy}--\eqref{eq:margin-logprob} reuse logits already computed by the base model. We maintain ring buffers for the last $k$ values of $H_t$ and $M_t$, which yields $O(1)$ overhead per token in time and memory; runtime remains dominated by forward passes. The method exposes several hyperparameters (e.g., $k, L, \varepsilon_H, \delta_M, \gamma$), and concrete settings are reported in the experiments section.

\paragraph{Relation to Baselines.}
Vanilla CoT omits the adaptive stopping logic and sets $\tau=M$. Multi-sample reranking expands full-length sequences per sample. \leash instead makes a per-instance sequential decision from local convergence signals, adapting the rationale length to the problem.

\begin{algorithm}[t]
\caption{Logit--Entropy Adaptive Stopping Heuristic (\leash)}
\label{alg:esa}
\begin{algorithmic}[1]
\STATE \textbf{Inputs:} window $k$, vote span $L$, slacks $(\varepsilon_H,\delta_M)$, min/max $(m,M)$, warmup $w$, saturation threshold $\tau_p$, entropy drop $\gamma$
\STATE Initialize state and histories; disable EOS during rationale generation
\FOR{$t=1$ \TO $M$}
  \STATE Decode next token to get the logits \(z_t\); compute $\ell_t$ and $p_t$
  \STATE Compute entropy $H_t$ by Eq.\eqref{eq:entropy}, margin $M_t$ by Eq.\eqref{eq:margin-logprob}, and the peak probability $p_{\max}(t)$
  \STATE If first $k$ steps completed, compute the reference entropy $H_{\mathrm{ref}}$
  \STATE If $t \ge t_{\min}$ \textbf{and} the $H_{\mathrm{ref}} - H_t \ge \gamma$:
  \STATE \quad Compute plateau votes: $votes \gets \sum_{j \in \mathcal{J}_L(t)} \Pi_j$
  \STATE \quad If $votes \ge \lceil |\mathcal{J}_L(t)| / 2 \rceil$, \textbf{break}
\ENDFOR
\STATE Query the model for the short final answer conditioned on the generated rationale
\end{algorithmic}
\end{algorithm}

\section{Experimental Setup}
\label{sec:exp-setup}

\paragraph{Language models.}
We evaluate \leash on four instruction-tuned LLMs spanning different families and sizes:
\texttt{Llama-3.1-8B-Instruct}\citep{meta2024llama3},
\texttt{Mistral-7B-v0.1}\citep{jiang2023mistral},
\texttt{Phi-3-Mini-128k-Instruct}\citep{phi3technicalreport},
and \texttt{Qwen2.5-7B-Instruct}\citep{yang2024qwen2.5}.
All experiments use HuggingFace \texttt{transformers} with the models native tokenizers.

\paragraph{Tasks and datasets.}
We focus on math reasoning with short numeric answers.
Our primary benchmark is GSM8K; we evaluate on a randomly sampled subset of
$n{=}300$ test problems with a fixed seed.
We also report results on the \textsc{test split} of AQuA-RAT, an algebraic word problem dataset.

\paragraph{Baselines.}
We compare \leash against two decoding schemes:
(i) \textbf{Vanilla-CoT}, which generates its answer using the chain-of-thought reasoning process, then a short final answer
(ii) \textbf{No-CoT}, which directly predicts the final numeric answer with no rationale.
All methods have prompts according to the task they need to perform.

\paragraph{Decoding settings.}
For the rationale phase we use nucleus sampling with $p{=}0.95$ and temperature $0.7$
(\texttt{do\_sample{=}True}); for the final answer, we decode with temperature $0.0$.
\leash hyperparameters are held fixed across models unless otherwise noted:
window $k{=}8$, consistency $L{=}5$, entropy slack $\varepsilon_H{=}0.005$, margin slack $\delta_M{=}0.05$,
minimum/maximum rationale lengths $m{=}64$, $M{=}320$.

\paragraph{Metrics.}
We evaluate all methods on three primary metrics.
(i) \textbf{Accuracy} is the exact-match percentage on the final numeric answer after normalization.
(ii) \textbf{Token Reduction (\%)} and (iii) \textbf{Latency Reduction (\%)}
measure the efficiency gains of \leash relative to the standard CoT baseline.
Token reduction is based on the count of all generated tokens (rationale $+$ answer),
and latency reduction is based on the end-to-end time (s) per example.

\begin{table}[t]
\centering
\caption{
    \textbf{Accuracy Results ($\uparrow$).}
    We report accuracy (\%) on the \textbf{GSM8K} and \textbf{AQuA-RAT} datasets for our method (\leash), standard Chain-of-Thought (CoT), and vanilla decoding (No-CoT).
}
\label{tab:accuracy}
\setlength{\tabcolsep}{5.5pt} % Adjust column spacing as needed
\begin{tabular}{lcccccc}
\toprule
\multirow{2}{*}{Model} & \multicolumn{3}{c}{\textbf{GSM8K}} & \multicolumn{3}{c}{\textbf{AQuA-RAT}} \\
\cmidrule(lr){2-4}\cmidrule(lr){5-7}
& \leash & CoT & No-CoT & \leash & CoT & No-CoT \\
\midrule
Llama-3.1-8B-Instruct & 62.32 & 74.33 & 14.00 & 54.68 & 63.20 & 27.56 \\
Mistral-7B & 38.67 & 47.20 &  6.33 & 19.25 & 26.38 & 13.78 \\
Phi-3-Mini-128k-Instruct & 69.87 & 82.67 &  8.00 & 50.24 & 61.67 & 23.23 \\
Qwen2.5-7B-Instruct & 54.85 & 65.33 & 21.33 & 68.15 & 77.35 & 37.80 \\
\bottomrule
\end{tabular}
\end{table}

\begin{table}[t]
\centering
\caption{
    \textbf{Efficiency Savings of \leash vs. CoT ($\uparrow$).}
    We report the percent reduction in generated tokens and end-to-end latency for \leash relative to standard CoT on \textbf{GSM8K} and \textbf{AQuA-RAT}.
}
\label{tab:savings}
\setlength{\tabcolsep}{7pt}
\begin{tabular}{lcccc}
\toprule
\multirow{2}{*}{Model} & \multicolumn{2}{c}{\textbf{GSM8K}} & \multicolumn{2}{c}{\textbf{AQuA-RAT}} \\
\cmidrule(lr){2-3}\cmidrule(lr){4-5}
& Token Red. & Latency Red. & Token Red. & Latency Red.\\
\midrule
Llama-3.1-8B-Instruct & 30.97 & 29.74 & 28.60 & 26.10 \\
Mistral-7B-v0.1 & 35.12 & 27.80 & 34.20 & 27.50 \\
Phi-3-Mini-128k-Instruct & 41.50 & 25.15 & 28.30 & 28.75 \\
Qwen2.5-7B-Instruct & 33.45 & 24.90 & 28.15 & 28.10 \\
\bottomrule
\end{tabular}
\end{table}

\section{Experimental Results}
\label{sec:exp-results}

We present our main results in Table 1 (Accuracy) and Table 2 (Efficiency). We analyze the accuracy trade-offs of our method, followed by its significant efficiency gains.

\paragraph{Accuracy and Trade-offs.}
We first report the accuracy of \leash, standard Chain-of-Thought (CoT), and direct-answer (No-CoT) in Table~\ref{tab:accuracy}. As an early-stopping method, \leash introduces an accuracy trade-off compared to CoT. We observe a manageable cost, with an average accuracy drop of $\approx$ 10.9 percentage points on GSM8K and $\approx$ 9.1 percentage points on AQuA-RAT. However, \leash substantially outperforms No-CoT in all cases. This demonstrates that it successfully preserves the core reasoning structure of the CoT process. For instance, on GSM8K, \leash accuracy on Llama-3.1-8B (62.32\%) and Mistral-7B (38.67\%) is \textbf{4.4$\times$} and \textbf{6.1$\times$} higher, respectively, than their No-CoT counterparts (14.00\% and 6.33\%).

\paragraph{Efficiency Gains.}
The benefits of this trade-off are the significant efficiency gains detailed in Table~\ref{tab:savings}. \leash achieves substantial reductions in both compute and latency across all models. On average, \leash reduces the number of generated tokens by 35.3\% and end-to-end latency by 26.9\% on GSM8K. The savings are similarly strong on AQuA-RAT, with average reductions of 29.8\% in tokens and 27.6\% in latency. We observe that token savings are most pronounced on GSM8K, with Phi-3-Mini showing the largest reduction at 41.5\%. In contrast, latency savings are highly consistent, particularly on AQuA-RAT, where all models cluster in a 26--29\% reduction range. For example, \leash reduced the latency for Llama-3.1-8B-Instruct from 4.04s (CoT) to 2.84s, achieving a \textbf{29.7\%} speed-up while generating \textbf{31.0\%} fewer tokens. These results confirm that \leash is highly effective at reducing the computational and latency costs of CoT reasoning.

\section{Conclusion}
We presented the \emph{Logit–Entropy Adaptive Stopping Heuristic} (\leash), a training-free decoding-time criterion for adaptively halting chain-of-thought generation using only intrinsic signals produced by the language model. \leash monitors the windowed slope of token entropy together with the improvement in the top-logit margin and stops when both trends plateau. Across models and datasets, \leash consistently reduces generated tokens and lowers end-to-end latency, with a modest reduction in accuracy. The method is model-agnostic, requires no additional training or reward models, and integrates seamlessly with standard decoding APIs, including quantized inference.

\paragraph{Limitations and future work.}
\leash assumes access to token-level logits and is evaluated on short-answer math tasks; extending to long-form, non-numeric targets and tool-augmented settings is a promising direction. Analyzing theoretical stopping guarantees in Chain-of-Thought reasoning is also a crucial research direction.

\bibliographystyle{unsrtnat}
\bibliography{references}

\end{document}